\documentclass[10pt,conference]{IEEEconf} 

\IEEEoverridecommandlockouts

\usepackage{graphics}
\usepackage{graphicx}
\usepackage{cite}
\usepackage{amsmath}
\usepackage{amssymb}
\usepackage{booktabs}
\usepackage{multirow}
\usepackage[dvipsnames]{xcolor}
\usepackage{comment}
\usepackage{ulem}

\title{\LARGE \bf
  MA-VLCM: A Vision Language Critic Model for Value Estimation of Policies in Multi-Agent Team Settings
}

\author{Shahil Shaik$^{1*}$, Aditya Parameshwaran$^{1*}$, Anshul Nayak$^{1}$, Jonathon M. Smereka$^{2}$, and Yue Wang$^{1}$
\thanks{This work was supported by the Automotive Research Center (ARC) at the University of Michigan, Ann Arbor, under Cooperative Agreement W56HZV-24-2-0001 with the US Army DEVCOM Ground Vehicle Systems Center (GVSC). Distribution Statement A. Approved for public release; distribution is unlimited. OPSEC \#10483}
  \thanks{$*$ Equal contributions}%
  \thanks{$^{1}$Shahil Shaik, Aditya Parameshwaran, Anshul Nayak and Yue Wang are with the Mechanical Engineering
    Department, Clemson University.}%
  \thanks{$^{2}$Jonathon M. Smereka is with the Ground
    Vehicle Systems Center U.S. Army CCDC}%
}

\begin{document}

\maketitle

\begin{abstract}
Multi-agent reinforcement learning (MARL) commonly relies on a centralized critic to estimate the value function. However, learning such a critic from scratch is highly sample-inefficient and often lacks generalization across environments. At the same time, large vision-language-action models (VLAs) trained on internet-scale data exhibit strong multimodal reasoning and zero-shot generalization capabilities, yet directly deploying them for robotic execution remains computationally prohibitive, particularly in heterogeneous multi-robot systems with diverse embodiments and resource constraints.
To address these challenges, we propose Multi-Agent Vision-Language-Critic Models (MA-VLCM), a framework that replaces the learned centralized critic in MARL with a pretrained vision-language model fine-tuned to evaluate multi-agent behavior. MA-VLCM acts as a centralized critic conditioned on natural language task descriptions, visual trajectory observations, and structured multi-agent state information. By eliminating critic learning during policy optimization, our approach significantly improves sample efficiency while producing compact execution policies suitable for deployment on resource-constrained robots. Results show good zero-shot return estimation on  models with differing VLM backbones on in-distribution and out-of-distribution scenarios in multi-agent team settings.
\end{abstract}

\section{INTRODUCTION}

Vision-Language-Action models (VLAs) have recently emerged as a powerful paradigm for controlling embodied robotic systems across a broad range of tasks. Unlike traditional reinforcement learning (RL) policies, VLAs demonstrate strong generalization to novel task specifications by leveraging a semantic understanding of both the task and the surrounding environment, such as $\pi_0$ \cite{intelligence2025pi_}. This capability is largely enabled by large vision-language model (VLM) backbones trained on internet-scale data, which are subsequently fine-tuned on diverse robotic datasets to produce action outputs.

Despite their effectiveness, modern VLAs suffer from slow inference speeds and large model sizes \cite{kim24openvla}, which significantly hinder their deployment in real-world robotic systems. This limitation becomes particularly pronounced in multi-agent settings, where practical systems often consist of heterogeneous teams of robots with varying physical dimensions, sensing modalities, actuation capabilities, and onboard computational resources \cite{ramachandran2021resilient}. Smaller or resource-constrained robots may lack the computational capacity to run a VLA onboard, and training a unified VLA across multiple robotic platforms is highly data-intensive, typically requiring costly retraining to learn platform-specific control dynamics.

In parallel, Multi-Agent Reinforcement Learning (MARL) has been a principled approach for learning joint policies for teams of agents under shared or partial observations \cite{lowe2017multi}. A widely adopted paradigm within MARL is centralized training with decentralized execution (CTDE), where a centralized critic leverages global information during training while execution policies remain decentralized and lightweight \cite{lowe2017multi, yu2021surprising}. However, MARL also suffers from well-known limitations: learning effective value functions in high-dimensional observation spaces is sample inefficient, sparse rewards make exploration difficult, and policies struggle to generalize across task specifications and environments.

To address these challenges, we propose the Multi-Agent Vision-Language-Critic Model (MA-VLCM) framework. Instead of directly deploying a VLA as an execution policy, the MA-VLCM is trained to evaluate and critique team-level robot behaviors. Specifically, the critic takes as input a textual task description, visual observations in the form of trajectory videos, and structured multi-agent state observations, evaluating the robot team's performance with respect to the task. Trained contrastively to assign higher values to behaviors that better satisfy team objectives, the VLCM acquires cross-task and cross-environment generalization. Our work represents a first step towards a generalized, sample-efficient multi-agent critic, eliminating the need to learn a critic from scratch during MARL policy training.

The main contributions of this work are:
\begin{itemize}
    \item We introduce MA-VLCM, a novel vision-language-based centralized critic that evaluates multi-agent robot team behavior conditioned on task instructions, visual trajectory data, and multi-agent observations.
    \item We demonstrate that the resulting critic model achieves strong task performance across multiple structured and unstructured multi-agent datasets, and on out of distribution task assignments as well.
\end{itemize}

\section{Related Work}

\subsection{VLM/VLA}

Foundational models such as large language models \cite{driess2023palm} and vision-language models \cite{liu2023visual} have transformed embodied AI. Pretrained robotic Vision-Language-Action (VLA) systems such as OpenVLA \cite{kim24openvla} demonstrate strong multi-task generalization by leveraging internet-scale data. However, VLA models face key limitations: they are computationally expensive and slow at inference \cite{zitkovich2023rt}, and adapting them across heterogeneous agents requires large-scale finetuning on extensive datasets \cite{bousmalis2023robocat}. Recent works have used foundation models for planning, reward modeling, and policy evaluation \cite{xie2023text2reward}, but these approaches are largely confined to single-agent settings and do not address centralized value estimation in multi-agent reinforcement learning.

\subsection{MARL}

Multi-Agent Reinforcement Learning (MARL) provides a principled framework for learning collaborative and competitive behaviors among interacting agents \cite{oroojlooy2023review}. Under the CTDE paradigm \cite{amato2024introduction}, algorithms such as MADDPG \cite{lowe2017multi} and MAPPO \cite{yu2022surprising} employ centralized critics to mitigate non-stationarity. However, centralized critics are sample inefficient in high-dimensional spaces, rarely generalize across task variations \cite{cobbe2019quantifying}, and lack semantic grounding in language or visual abstractions. Recent efforts improve sample efficiency via representation learning \cite{vadakkepat2024mapolso} or contrastive objectives over trajectories \cite{liu2023cia}, but still learn the critic within the RL loop and do not leverage pretrained multimodal foundation models as generalized value estimators.

\subsection{Vision Language Critics}

Recent works have explored leveraging vision-language models as critics or reward models for robotic RL. VLAC \cite{zhai2025vision} unifies actor and critic roles within a single autoregressive architecture, while VLC \cite{alakuijala2024video} and LIV \cite{ma2023liv} learn transferable reward functions via contrastive video-language alignment or temporal ranking objectives. While these methods demonstrate improved sample efficiency in real-world manipulation, they are predominantly designed for single-agent settings. In contrast, our work extends the vision-language critic paradigm to a multi-agent centralized setting, where a pretrained VLM serves as a coordination-aware critic that evaluates team-level behavior and supports decentralized multi-agent policy learning.

\begin{figure*}[ht]
    \centering
    \includegraphics[width=\textwidth]{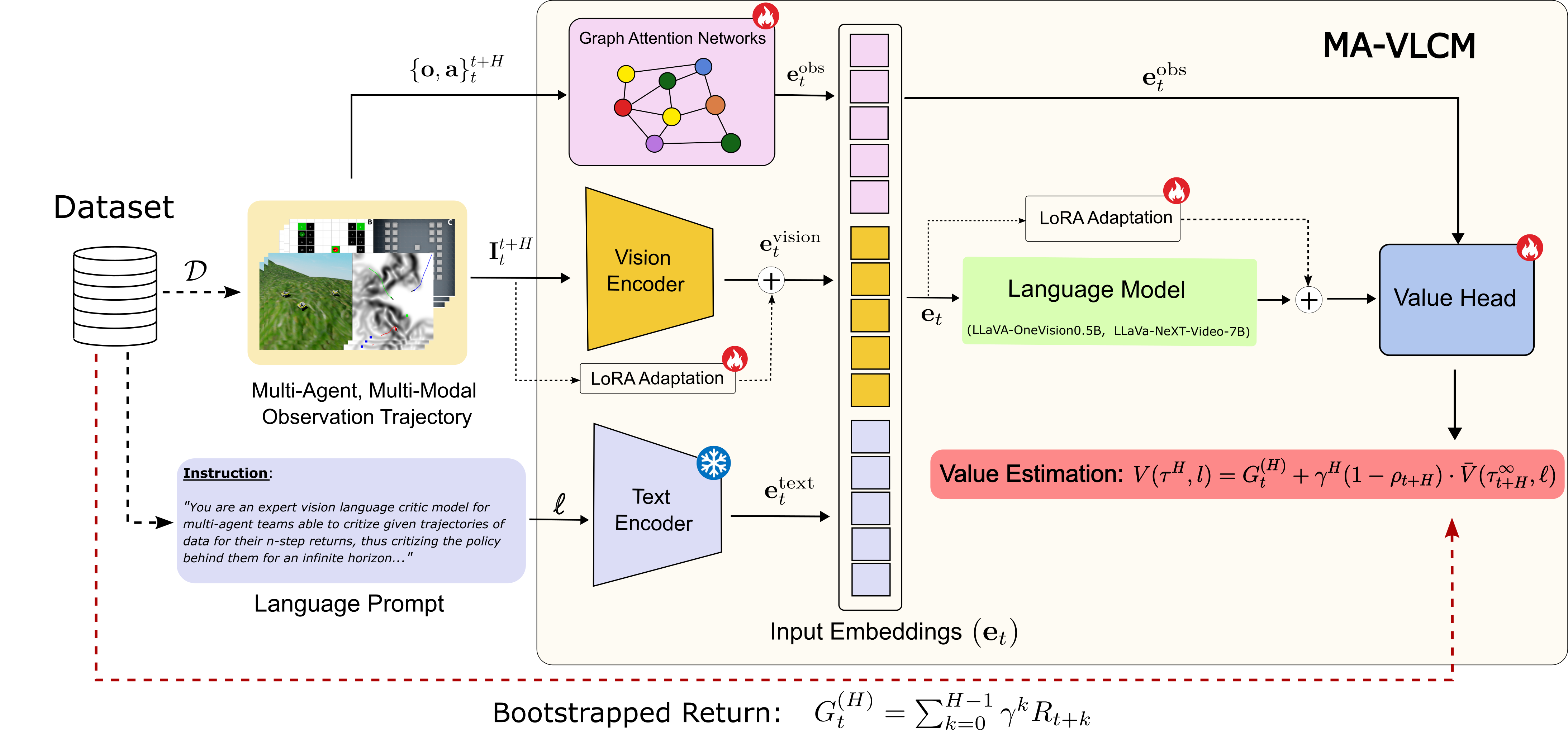}
    \caption{Overall architecture of the MA-VLCM training (\textcolor{red}{- -}) and inference (- -) directions. The MA-VLCM dataset ($\mathcal{D}$) contains multi-modal, multi-agent dataset that is used to send trajectories of vision $\mathbf{I}_t$, textual $\ell$ and agent observation $\mathbf{o_t}$ and actions $\mathbf{a}_t$ data to train the VLCM. The main components that are trained are the GAT, Value head and LoRA adaptors on the vision encoder and language model. The output embeddings from the the encoders $\mathbf{e}_t$ and the GAT are concatentated before sending them to the language model. The output of the value head would be the estimated long-term return of the trajectory of data along with the language prompt sent during inference.}
    \label{fig:approach}
\end{figure*}

\section{Preliminaries}
\subsection{Problem Formulation}

We consider a cooperative MARL problem modeled as a Dec-POMDP
$
\langle \mathcal{N}, \mathcal{T}, \{\mathcal{O}^i\}_{i=1}^n,
\{\mathcal{A}^i\}_{i=1}^n, R, P, \gamma \rangle .
$
Here, $\mathcal{N}=\{1,\dots,n\}$ denotes the set of agents and $\mathcal{T}$ the task set. 
Each agent $i$ has observation and action spaces 
$\mathcal{O}^i \subset \mathbb{R}^p$ and 
$\mathcal{A}^i \subset \mathbb{R}^q$, with joint spaces 
$\mathcal{O}=\prod_{i=1}^n \mathcal{O}^i$ and 
$\mathcal{A}=\prod_{i=1}^n \mathcal{A}^i$. 
The transition kernel 
$P:\mathcal{O}\times\mathcal{A}\times\mathcal{O}\to[0,1]$ 
defines the environment dynamics, 
$R:\mathcal{O}\times\mathcal{A}\to[-R_{\min},R_{\max}]$ 
is the VLCM-compatible reward (see Section~\ref{section:VLCM}), 
and $\gamma\in[0,1)$ is the discount factor.

Agent interactions are modeled as a dynamic graph 
$G=(\mathcal{N},\mathcal{E})$, where edges represent time-varying communication links. 
During training, the centralized VLCM processes graph-structured information, 
while policies execute using only local observations $o_t^i$ under the CTDE paradigm. 
Each agent selects actions $a_t^i \sim \pi^i$, forming the joint policy 
$\pi=\prod_{i=1}^n \pi^i$. 
Together with $P$, this induces the discounted state distribution
$
\rho_{\pi}(\cdot)=\sum_{t=0}^{\infty}\gamma^t 
\Pr(\mathbf{o}_t \mid \pi)
$
\cite{wen2022multi}. 
Agents receive a shared team reward
$
R(\mathbf{o}_t,\mathbf{a}_t)
= \frac{1}{n}\sum_{i\in\mathcal{N}} R^i(o_t^i,a_t^i),
$
and observe the next local observations $o_{t+1}^i$.

We consider the fully cooperative setting in which all agents optimize a shared team reward. The goal is to learn local policies $\{\pi^i\}_{i=1}^n$ that maximize the expected discounted team return:
\begin{equation}\label{eq:objective_function}
    J\left(\pi\right) = \mathbb{E}_{\pi}\left[\sum_{t=0}^{\infty}\gamma^t R(\mathbf{o}_t,\mathbf{a}_t)\right]
\end{equation}

\subsection{Graph Attention Networks}

Graph Attention Networks (GATs) \cite{velivckovic2017graph, brody2021attentive} learn expressive representations from graph-structured data through attention-based message passing. At time step $t$, each node is associated with a feature vector $\mathbf{h}_t^i \in \mathbb{R}^{d}$. For a pair of neighboring nodes $(i,j)$ in graph $G$, the unnormalized attention score is computed as
\begin{equation}
e(\mathbf{h}_t^i,\mathbf{h}_t^j) =
\mathrm{LeakyReLU}
\!\left(
\mathbf{q}^{\top}
\left[
\mathbf{W}\mathbf{h}_t^i
\|
\mathbf{W}\mathbf{h}_t^j
\right]
\right),
\label{eq:attention_score}
\end{equation}
where $\mathbf{W} \in \mathbb{R}^{d' \times d}$ is a learnable linear projection, 
$\mathbf{q} \in \mathbb{R}^{2d'}$ is a trainable attention vector, and $\|$ denotes vector concatenation.

The attention coefficients are obtained via softmax normalization over the neighborhood:
\begin{equation}
\alpha_{ij,t}
=
\frac{
\exp\!\left(e(\mathbf{h}_t^i,\mathbf{h}_t^j)\right)
}{
\sum_{j'\in\mathcal{N}^i}
\exp\!\left(e(\mathbf{h}_t^i,\mathbf{h}_t^{j'})\right)
}.
\label{eq:norm_attention_score}
\end{equation}

The updated node representation is then
\begin{equation}
\mathbf{u}_t^i =
\sigma
\!\left(
\sum_{j\in\mathcal{N}^i}
\alpha_{ij,t}\,
\mathbf{W}\mathbf{h}_t^j
\right),
\end{equation}
where $\sigma(\cdot)$ denotes a nonlinear activation function.


\section{Methodology}
\label{section:VLCM}

\subsection{Multi-Agent Vision-Language-Critic Model}
The primary role of a centralized critic, $V:\mathcal{O}\rightarrow\mathbb{R}$, in MARL is to estimate the long-term return of a team of agents executing a joint policy $\pi$ from a global observation $\mathbf{o}\in\mathcal{O}$. Although centralized critics improve stability under CTDE, they are typically learned from scratch for each task distribution $\mathcal{T}$, resulting in high sample complexity and limited cross-task generalization. We propose the MA-VLCM (Fig~\ref{fig:approach}), which replaces a task-specific critic with a foundation-model-based critic that leverages (i) large-scale pre-trained vision-language priors and (ii) structured multi-agent temporal reasoning via graph attention. The key idea is to reinterpret value estimation as a multi-modal sequence, where the model jointly reasons over the visual trajectory, a natural-language task specification, and structured multi-agent observations.

\subsubsection{Multi-modal conditioning with observation tokens.}
Let a trajectory segment under policy $\pi$ be
\begin{equation}\label{eq:trajectory}
\tau^{t+H}_{t} = \{\mathbf{o},\mathbf{I},\mathbf{a}\}_{t}^{t+H}
\end{equation}
where $\mathbf{o}_k$, $\mathbf{I}_k$, and $\mathbf{a}_k$ denote the multi-agent state observations, corresponding bird's eye view (BEV) video frame, and actions at timestep $k$; let $\ell$ denote the language task prompt. For simplicity, we denote $\tau_{t}^{t+H}$ as $\tau^H$. We encode the structured robotic observations at each timestep $t$ using temporal graph-attention (GAT) module,
\begin{equation}
\mathbf{h}_t = f_{\mathrm{GAT}}\{\mathbf{o}, \mathbf{a}\}_{t}^{t+H}
\end{equation}
which produces permutation-invariant embeddings that capture inter-agent interactions. These embeddings are projected into the token embedding space of the vision-language backbone:
\begin{equation}
\mathbf{e}^{\mathrm{obs}}_t = \mathbf{W}_{\mathrm{proj}}\,\mathbf{h}_t .
\end{equation}
We introduce a dedicated learnable observation token $\langle\texttt{OBS}\rangle$ and instantiate time-indexed tokens $\langle\texttt{OBS}\rangle_t$ whose input embeddings are set to $\mathbf{e}^{\mathrm{obs}}_t$. The transformer input is formed by concatenating the video ($\mathbf{e}_t^\text{vision}$), text ($\mathbf{e}_t^\text{text}$), and GAT output embeddings ($\mathbf{e}_t^\text{obs}$) as:
\begin{equation*}
    \mathbf{e}^t = [\mathbf{e}_t^\text{obs}, \mathbf{e}_t^\text{vision},\mathbf{e}_t^\text{vision}]
\end{equation*}
The concatentated embedding space $\mathbf{e}^t$ is then sent as input to a backbone language model, that is further processed to send the outputs to a value head. This enables the backbone to jointly capture spatial-temporal visual features, task-level semantic intent, and explicit robotic state encoded by the graph module. We also send the observation embeddings from the GAT to the value head directly as well as a skip-layer connection to retain information during backpropagation through the large language model.

\subsubsection{Policy-space interpretation of the MA-VLCM}

Let $\Phi(\tau^{H},\ell)\in\mathbb{R}^m$ denote the latent representation produced by the vision-language backbone after cross-modal reasoning. We interpret $\Phi(\tau^{H},\ell)$ as an implicit embedding of the executed joint policy $\pi\in\Pi$ under task $\ell$. Let $d:\mathbb{R}^m\times\mathbb{R}^m\rightarrow\mathbb{R}_{\ge 0}$ be a metric on the latent space, and let $\Pi^\star_\ell\subset\Pi$ denote the set of desirable policies for task $\ell$. Define the corresponding set of desirable latent embeddings
\begin{equation}
\mathcal{Z}^\star_\ell \;:=\; \big\{\, \Phi(\tau^{H^{\star}},\ell)\in\mathbb{R}^m \;\big|\; \tau^{H^\star} \sim \pi^\star,\ \pi^\star\in\Pi^\star_\ell \,\big\}.
\label{eq:optimal_latent_set}
\end{equation}
We seek to maximize the separation between sub-optimal behaviors and the desirable set by optimizing the backbone parameters $\psi$:
\begin{equation}
\max_{\psi}\; d\!\left(\Phi(\tau^{H},\ell),\mathcal{Z}^\star_\ell\right)
\;:=\;
\max_{\psi}\; \inf_{\mathbf{z}^\star\in\mathcal{Z}^\star_\ell}
d\!\left(\Phi(\tau^{H},\ell),\mathbf{z}^\star\right),
\label{eq:policy_distance_value}
\end{equation}
where the contrastive loss ($\mathcal{L}_{con}$) is induced by the metric $d$ on $\mathbb{R}^m$. We optimize the backbone using a contrastive objective that treats embeddings of desirable trajectories as positives and sub-optimal trajectories as negatives, directly enforcing the margin defined in~\eqref{eq:policy_distance_value}.

The final scalar value is obtained via a lightweight linear head:
\begin{equation}
V^\pi(\tau^{H},\ell) = \mathbf{w}^\top \Phi(\tau^{H},\ell) + b,
\end{equation}
where $\mathbf{w}\in\mathbb{R}^m$ and $b\in\mathbb{R}$ are trainable parameters. The value head is trained using standard temporal-difference or Monte-Carlo regression targets,
\begin{equation}
\mathcal{L}_{\mathrm{value}} 
=
\left( V^\pi(\tau^{H},\ell) - \hat{R} \right)^2,
\end{equation}
with $\hat{R}$ denoting the TD or return estimate. The overall optimization combines representation shaping and value calibration:
\begin{equation}
\mathcal{L}_{\mathrm{total}} 
=
\mathcal{L}_{\mathrm{value}} 
+ \lambda \mathcal{L}_{\mathrm{con}},
\end{equation}
where $\lambda$ balances latent separation and return regression.

Under this formulation, the backbone learns a structured latent space: trajectories from high-performing policies cluster together, while sub-optimal behaviors lie farther away. The value head maps this structured representation to a calibrated scalar estimate. By conditioning on observation tokens, the learned representation remains grounded in multi-agent robotic state dynamics, ensuring semantically meaningful and physically consistent value estimates.

\subsection{Multi-Agent Vision Language Critic Dataset}
\label{subsec:dataset}

A core contribution of our work is a comprehensive multi-agent, multi-modal dataset designed to train the VLCM. Prior works have focused on single agent offroad navigation \cite{triest2022tartandrive} or manipulation tasks \cite{Khazatskyetal2024}, comprising only optimal data. Our data collection strategy captures multi-agent team tasks with coordination aspects for both structured and unstructured environments.
Formally, we define $\mathcal{D} = \{ \tau_i \}_{i=1}^N$ consisting of $N$ trajectories. Each trajectory $\tau^{H}$ of horizon $H$ is a sequence of tuples $\tau = \{\mathbf{o},\mathbf{I}, \mathbf{a}\}_{t}^{t+H}$ conditioned on a global language prompt $\ell$.

\textbf{Visual Observations ($\mathbf{I}_t, \tilde{\mathbf{I}}_t$):} 
We collect multi-modal Bird's Eye View (BEV) visual inputs 
$\mathbf{I}_t \in \mathbb{R}^{n \times h \times w \times c}$ for $n$ agents. These modalities may include RGB imagery, grayscale terrain traversability maps, or rasterized semantic representations. The BEV can also be rasterized to produce a semantic map $\tilde{\mathbf{I}}_t$ from state observations. Examples are shown in Figures~\ref{fig:rware_image} and~\ref{fig:offroad_image}.

\textbf{State Observations ($\mathbf{o}_t$):} Proprioceptive state matrix $\mathbf{o}_t \in \mathbb{R}^{n \times p}$ including pose, velocity, orientation and actions of every agent. To support unified VLCM training across environments with different observation dimensions, we mask the corresponding GAT input nodes accordingly.
    
\textbf{Graph Topology ($\mathbf{A}_t$):} Time-varying multi-agent communication encoded as an adjacency matrix $\mathbf{A}_t \in \{0,1\}^{n \times n}$, enabling the VLCM to model connectivity and information flow while remaining robust to varying team configurations.
    
\textbf{Language Task Prompts ($\ell$):} Each trajectory is associated with a caption $\ell$ describing the task and environment configuration, providing richer expressivity than one-hot task IDs. For example, in RWARE, $\ell$ specifies: \begin{quote}
\textit{``Two agents must coordinate to pick two boxes and deliver them to the goal locations''} for (\texttt{rware-tiny-2ag}) vs. \textit{``Four agents must coordinate to pick two boxes...''} for (\texttt{rware-tiny-4ag})
\end{quote}
This linguistic grounding enables the critic to evaluate policies relative to semantic instructions. An example prompt is shown in Figure~\ref{fig:approach}.

\textbf{Trajectory Returns ($R_t$):} The MA-VLCM predicts the expected infinite-horizon discounted return given a trajectory segment. We compute a ground-truth reward matrix $R \in \mathbb{R}^{n \times T}$ containing the dense rewards for each agent $i \in n$ at time $t \in T$. The MA-VLCM accepts a context window of $H$ frames as trajectory $\tau^H$ (Equation~\ref{eq:trajectory}) and computes an \textit{n-step discounted return}: \begin{equation} G_t^{(H)} = \sum_{k=0}^{H-1} \gamma^k R_{t+k} \end{equation} We bootstrap beyond the clip length using an exponential moving average (EMA) target network $\bar{V}$: 
\begin{equation} 
V(\tau^{H},\ell) = G_t^{(H)} + \gamma^{H} (1 - \rho_{t+H}) \cdot \bar{V}(\tau_{t+H}^{\infty}, \ell) 
\end{equation}
where $\rho_{t+H} \in \{0,1\}$ is the terminal flag. The EMA target parameters $\bar{\theta}$ are updated as:
\begin{equation}
    \bar{\theta} \leftarrow \alpha \cdot \bar{\theta} + (1 - \alpha) \cdot \theta
\end{equation}
where $\alpha \in [0, 1]$ is the smoothing coefficient (typically close to $1$). We limit the max number of steps to 64 after $t+H$. This mechanism mitigates instabilities inherent in training large-scale foundation models within an RL loop \cite{zhang2026ema}. The MA-VLCM is trained using mean-squared error combined with a contrastive pairwise loss that preserves the \textit{ranking} of trajectories by quality.

\subsubsection{Multi-Level Trajectory Optimality}
\label{subsec:trajectory_optimality}
Unlike large-scale VLA datasets that predominantly contain expert demonstrations, the MA-VLCM dataset captures coordination and team behaviours spanning multiple levels of optimality. Since the VLCM is trained as a critic, collecting data exclusively from high-reward policies would limit the critic's discriminative capacity.

For both environments, we sample trajectories from policies at varying stages of training, yielding a spectrum of cumulative returns from near-random to near-optimal. Let $\Pi = \{\pi_1, \pi_2, \dots, \pi_K\}$ denote $K$ policies with varying performance. For each $\pi_k \in \Pi$, we collect trajectories $\mathcal{D}_k = \{\tau_j \sim \pi_k\}_{j=1}^{M_k}$, where:
\begin{equation}
    V^{\pi_1} \leq V^{\pi_2} \leq \dots \leq V^{\pi_K}.
\end{equation}
The complete dataset $\mathcal{D} = \bigcup_{k=1}^{K} \mathcal{D}_k$ provides a holistic representation of the policy space, enabling fine-grained value distinctions across the full performance spectrum.

\section{Experiments}
We evaluate the proposed MA-VLCM framework across two distinct multi-agent domains: a structured robotic warehouse environment and an unstructured offroad navigation environment.

\subsection{Data Collection Pipeline}
\label{subsec:data_collection}
 \begin{figure}[h]
    \centering
    \includegraphics[width=0.9\columnwidth]{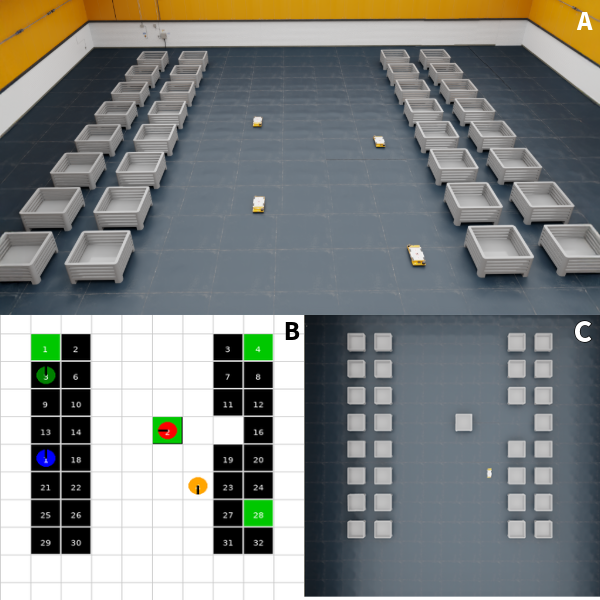}
    \caption{(a) The Robotic Warehouse (RWARE) 3D environment created in Isaac Sim. (b) A rendering of \texttt{rware-4ag} grid environment with 4 agents and 4 requested boxes (in green) describing the rasterized semantic image, (c) and its equivalent BEV camera image collected from the same environment shown in (a).}
    \label{fig:rware_image}
\end{figure} 
\subsubsection{Structured Environment: Robotic Warehouse (RWARE)}\label{sec:rware}
We utilize the RWARE~\cite{papoudakis2021benchmarking} environment (Figure~\ref{fig:rware_image}). Trajectories are sampled from multiple RWARE configurations (varying agent counts, shelf layouts, and task specifications) using a low-fidelity grid-world simulator. To obtain high-fidelity visual observations, we execute a \textit{sim-to-sim transfer}: trajectories are replayed in a photorealistic warehouse constructed in NVIDIA Isaac Sim. This pipeline decouples policy evaluation from visual fidelity, allowing the VLCM to perform inference in the high-dimensional observation space without retraining the underlying policies.
 \begin{figure}[!h]
    \centering
    \includegraphics[width=0.9\columnwidth]{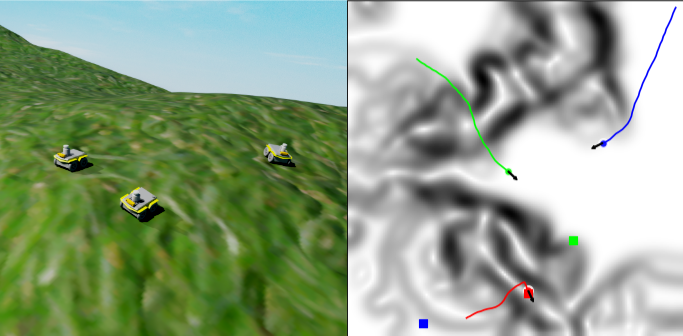}
    \caption{An unstructured offroad environment created in high-fidelity simulator with 3 Clearpath Jackal robots (left), and a corresponding rasterized top-down view (right), with the agents represented as color-coded markers moving to their target location on a traversability map generated from the high-fidelity simulator.}
    \label{fig:offroad_image}
\end{figure} 
\subsubsection{Unstructured Environment: Offroad Navigation}
We consider a multi-agent ground robot team operating in an offroad Isaac Sim environment (Figure~\ref{fig:offroad_image}). Each agent is tasked with a Go-To-Goal objective while avoiding collisions and selecting highly traversable paths. The BEV representation consists of rasterized blobs encoding each robot's position, trajectory from start, and goal location using color-coded markers, overlaid on an offroad traversability map. The traversability maps are generated by computing local gradients of terrain constructed in Isaac Sim. The environment is first voxelised to 0.05m voxel size, a 2.5D terrain height map is generated, and then converted to a traversability map based on roll and pitch limits with friction coefficients set to 0.7. Suboptimality is generated by adding noise to control commands, intentionally missing goals, and inducing collisions. A combination of offline path planning with weighted-Dijkstra* and online path planning with ORCA~\cite{berg2011reciprocal} is used to collect and replay trajectories via the same sim-to-sim pipeline as RWARE.

\subsection{Training Implementation}
\label{subsec:training_impl}
We train the MA-VLCM using two frozen VLM backbone variants: LLaVA-NeXT-Video-7B-32K and LLaVA-OneVision (OV)-Qwen2-0.5B, adapted via Low-Rank Adaptation (LoRA). Trajectories are serialized into WebDataset format for efficient distributed training across two NVIDIA H200 GPUs utilizing a combined temporal-difference and contrastive pairwise objective. Key hyperparameters are summarized in Table~\ref{tab:hyperparameters}.

\begin{table}[ht]
\centering
\caption{Key MA-VLCM Hyperparameters}
\label{tab:hyperparameters}
\renewcommand{\arraystretch}{1.2}
\resizebox{\columnwidth}{!}{%
\begin{tabular}{@{}ll@{}}
\toprule
\textbf{Parameter} & \textbf{Value} \\ 
\midrule
LoRA Rank ($r$), Scaling ($\alpha$) & $r=16$, $\alpha=32$  \\
Graph Attention Network (GAT) Dim & 256  \\
Optimizer & AdamW  \\
Learning Rate & $3 \times 10^{-4}$  \\
Effective Batch Size & 32  \\
Video Clip Length ($H$) & 32 frames  \\
\bottomrule
\end{tabular}%
}
\end{table}

\section{Results}
We present experimental results evaluating the MA-VLCM as a centralized critic, focusing on \textit{value estimation}. Provided a multi-agent trajectory segment, task description, and structured team observations, the model predicts the expected discounted return. Accurate value estimation is a prerequisite for using the VLCM as a replacement for a learned centralized critic in MARL.

\subsection{Value Estimation}
\label{subsec:value_estimation}
We evaluate against ground-truth discounted returns using frozen baselines and LoRA-augmented variants of both the VLM backbones (0.5B and 7B models). Models are evaluated under In-Distribution (ID) and Out-of-Distribution (OOD) domains. We report Spearman Rank Correlation ($\rho$), MSE/MAE, and mean prediction interval width (MPIW) over 100 test trajectory samples per configuration in Table~\ref{tab:value_estimation}.

\begin{table}[!h]
\centering
\caption{Value Estimation: In-Distribution vs.\ Out-of-Distribution}
\label{tab:value_estimation}
\renewcommand{\arraystretch}{1.2}
\resizebox{\columnwidth}{!}{%
\begin{tabular}{l|cc|cc}
\toprule
\multirow{2}{*}{\textbf{Metric}} & \multicolumn{2}{c|}{\textbf{In-Distribution}} & \multicolumn{2}{c}{\textbf{Out-of-Distribution}} \\
 & \textbf{RWARE} & \textbf{Offroad} & \textbf{RWARE} & \textbf{Offroad} \\
\midrule
\multicolumn{5}{l}{\textit{GNN + Value Head + LLaVA-0.5B-OV (Baseline)}} \\
\hspace{1em}seconds/iteration $\downarrow$ & \multicolumn{4}{c}{\textbf{0.20}} \\
\hspace{1em}Spearman $\rho$ $\uparrow$ & 0.80 & 0.82 & 0.78 & 0.86 \\
\hspace{1em}MSE $\downarrow$ & 6.11 & 17.71 & 13.24 & 22.08 \\
\hspace{1em}MAE $\downarrow$ & 5.10 & 2.04 & 5.22 & 3.03 \\
\hspace{1em}Pred.\ Interval Width & 2.60 & 3.24 & 2.82 & 3.74 \\
\midrule
\multicolumn{5}{l}{\textit{Baseline + LoRA (LLaVA-0.5B-OV)}} \\
\hspace{1em}seconds/iteration $\downarrow$ & \multicolumn{4}{c}{\textbf{0.23}} \\
\hspace{1em}Spearman $\rho$ $\uparrow$ & 0.95 & 0.96 & 0.86 & 0.93 \\
\hspace{1em}MSE $\downarrow$ & 1.68 & 15.40 & 2.92 & 25.83 \\
\hspace{1em}MAE $\downarrow$ & 1.04 & 3.50 & 1.37 & 4.31 \\
\hspace{1em}Pred.\ Interval Width & 3.52 & 7.21 & 5.77 & 10.15 \\
\midrule
\multicolumn{5}{l}{\textit{GNN + Value Head + LLaVA-7B (Baseline)}} \\
\hspace{1em}seconds/iteration $\downarrow$ & \multicolumn{4}{c}{\textbf{0.69}} \\
\hspace{1em}Spearman $\rho$ $\uparrow$ & 0.71 & 0.65 & 0.62 & 0.59 \\
\hspace{1em}MSE $\downarrow$ & 5.60 & 48.65 & 2.02 & 56.96 \\
\hspace{1em}MAE $\downarrow$ & 0.69 & 2.94 & 1.81 & 3.18 \\
\hspace{1em}Pred.\ Interval Width & 1.27 & 1.27 & 1.28 & 1.38 \\
\midrule
\multicolumn{5}{l}{\textit{Baseline + LoRA (LLaVA-7B)}} \\
\hspace{1em}seconds/iteration $\downarrow$ & \multicolumn{4}{c}{\textbf{0.72}} \\
\hspace{1em}Spearman $\rho$ $\uparrow$ & 0.84 & 0.72 & 0.80 & 0.70  \\
\hspace{1em}MSE $\downarrow$ & 1.61 & 35.68 & 3.22 & 47.39 \\
\hspace{1em}MAE $\downarrow$ & 0.95 & 10.27 & 1.52 & 11.47 \\
\hspace{1em}Pred.\ Interval Width & 1.47 & 1.29 & 1.58 & 0.79 \\
\bottomrule
\end{tabular}%
}
\end{table}

\begin{figure}[!h]
    \centering
    \includegraphics[width=0.9\columnwidth]{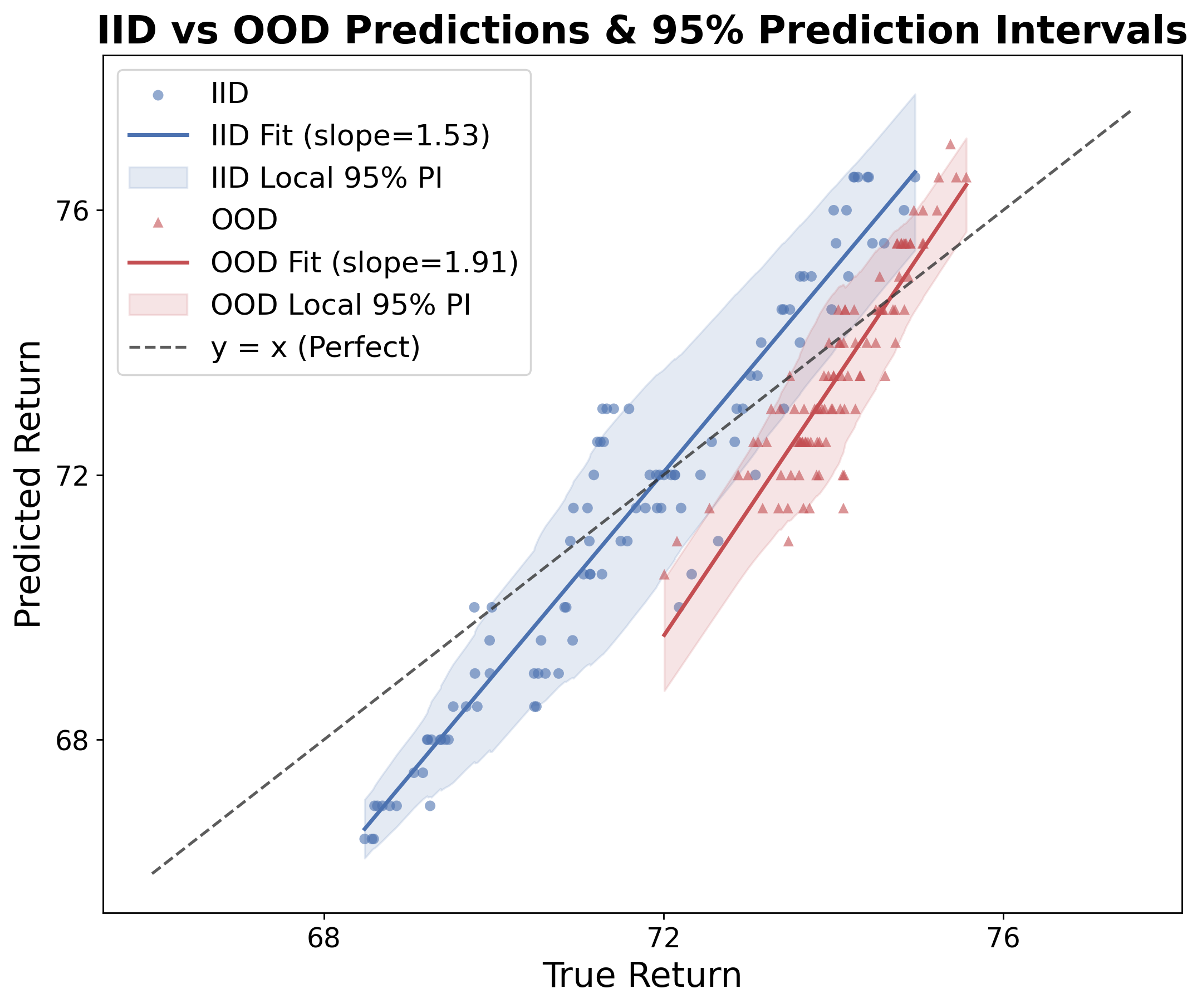}
    \caption{Value Estimation on IID vs OOD for RWARE environment with 0.5B VLCM}
    \label{fig:iid_ood_rware_0.5B}
\end{figure}

The results presented highlight the critical role of LoRA in utilizing the vision-language backbone for multi-agent value estimation. Integrating LoRA significantly enhances the model's ability to rank policies, which is the primary objective of the contrastive  framework. For instance, the LLaVA-0.5B-OV model augmented with LoRA achieves a significant improvement in Spearman rank correlation ($\rho$) across all evaluation settings, jumping from 0.80 to 0.95 in ID RWARE and from 0.86 to 0.93 in OOD Offroad as can be seen in Figure~\ref{fig:iid_ood_rware_0.5B}. This indicates that fine-tuning the vision and language weights is essential for the critic to correctly interpret novel task specifications and generalize to unseen environmental penalty structures. However, this improved spearman ranking capability sometimes trades off with absolute values, as evidenced by the elevated MSE in the OOD Offroad domain.  For the 0.5B model, when LoRA is applied the MSE (22.08 $\rightarrow$ 25.83) and the prediction interval widths (e.g., 3.74 $\rightarrow$ 10.15 in OOD Offroad) both increased suggesting the model's improved discriminative power comes at the cost of wider uncertainty bands.

\begin{figure}[!h]
    \centering
    \includegraphics[width=0.9\columnwidth]{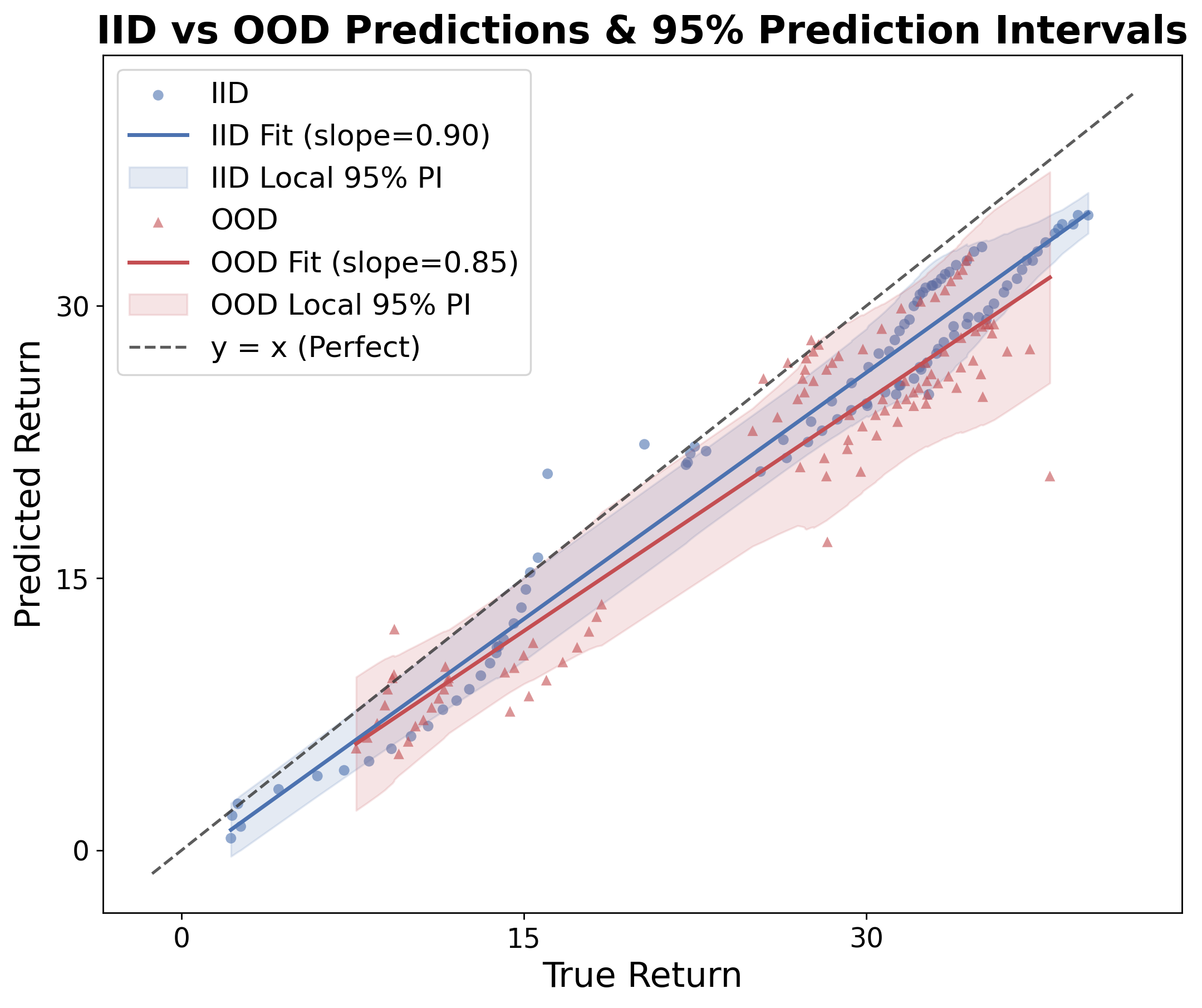}
    \caption{Value Estimation on IID vs OOD for Offroad environment with 0.5B VLCM}
    \label{fig:iid_ood_offroad_0.5B}
\end{figure}

\begin{figure}[!ht]
    \centering
    \includegraphics[width=\columnwidth]{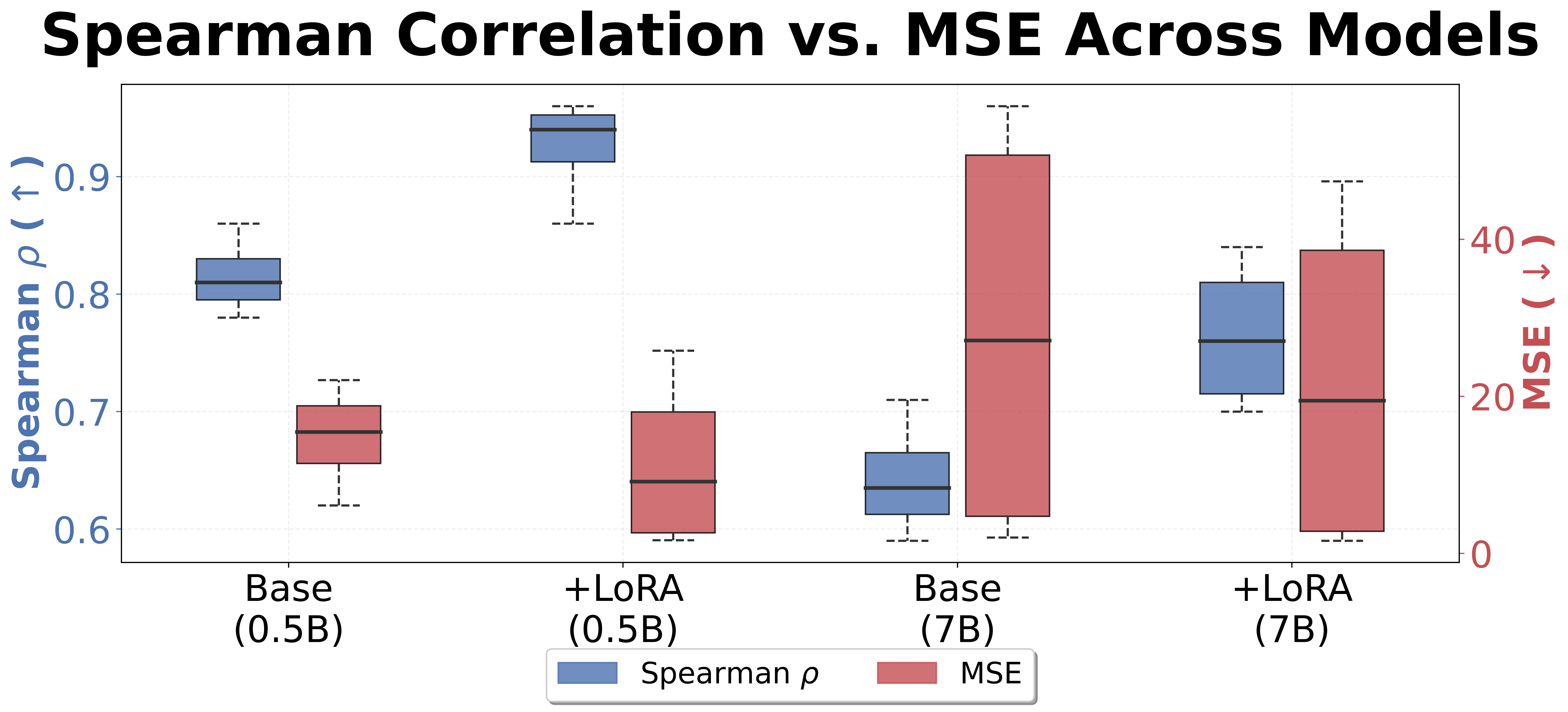}
    \caption{Box plot comparison for all 4 models on Spearman correlations and MSE values across both the environments}
    \label{fig:sample_comparison}
\end{figure}

Figure~\ref{fig:iid_ood_rware_0.5B} shows  that for the RWARE environment, the predicted returns near the average of the sampled test trajectories are close to the true average return. However, some predictions at the extreme high and low ends are incorrect. For Figure~\ref{fig:iid_ood_offroad_0.5B}, the predictions show a constant offset from the true return, but their slope is closer to the ideal line $y=x$ indicating a high Spearman correlation. This indicates that in the Offroad environment, MA-VLCM performs better in contrastive comparison than in MSE-based value estimation. The opposite trend is observed in the RWARE environment.
 Similarily, looking at the box plot shown in Figure~\ref{fig:sample_comparison}, we clearly see that the 0.5B parameter model outperforms the 7B model in both the Spearman correlation and the MSE errors.

 The 0.5B baseline model inferences trajectory clips at 0.20 seconds per iteration, whereas the 7B baseline requires 0.69 seconds; a \textbf{$3.45\times$} slowdown with no proportional gain in accuracy. Beyond inference speed, the LoRA-augmented 0.5B model achieves the highest Spearman $\rho$ across all settings (0.95, 0.96, 0.86, 0.93) , substantially outperforming the LoRA-augmented 7B variant (0.84, 0.72, 0.80, 0.70) as shown in Table.\ref{tab:value_estimation}. The relative underperformance of the 7B model is characterized by its elevated baseline MSE values (e.g., 48.65 for ID Offroad and 56.96 for OOD Offroad) and notably narrow prediction intervals ($\sim$1.27 to 1.39) that fail to capture the true variance. This suggests that the larger model's pre-trained latent space is overly rigid to adapt to multi-agent value regression. Without significantly larger dataset scales or extensive hyperparameter tuning, the 7B model struggles to adapt its complex representations to predict precise scalar return estimates, while the lighter 0.5B backbone proves more amenable to efficient fine-tuning.

\section{CONCLUSIONS}

In this work, we introduced MA-VLCM, a novel Vision-Language-Critic model designed to provide generalized value estimation for multi-agent reinforcement learning (MARL). Concurrently, it addresses the  sample inefficiency and limited task generalization associated with traditional MARL centralized critics. To train MA-VLCM, we generated a comprehensive multi-modal dataset encompassing both structured warehouse environment and unstructured offroad navigation tasks, purposefully incorporating trajectories spanning a wide spectrum of optimality. Our evaluations revealed that MA-VLCM robustly estimates policy returns and generalizes to challenging out-of-distribution scenarios, including varying agent configurations and novel constraints specified via natural language prompts. In future research, we intend to integrate the pre-trained MA-VLCM as a generalized centralized critic to train MARL policies in a few-shot approach for OOD settings. Furthermore, we plan to expand the MA-VLCM dataset by incorporating additional multi-agent datasets with heterogeneous teams.

\addtolength{\textheight}{-6cm}   



\bibliographystyle{IEEEtran}
\bibliography{References}

\end{document}